\documentclass[10pt,twocolumn,letterpaper]{article}

\usepackage{iccv}
\usepackage{times}
\usepackage{epsfig}
\usepackage{graphicx}
\usepackage{amsmath}
\usepackage{amssymb}

\usepackage{caption}
\usepackage{subcaption}

\usepackage[breaklinks=true,bookmarks=false]{hyperref}

\iccvfinalcopy % *** Uncomment this line for the final submission

\def\iccvPaperID{****} % *** Enter the ICCV Paper ID here

\ificcvfinal\fi

\begin{document}

%%%%%%%%% TITLE
\title{Adapting HouseDiffusion for conditional Floor Plan generation on Modified Swiss Dwellings dataset}

\author{Emanuel Kuhn\\
TU Delft\\
{\tt\small e.f.m.kuhn@student.tudelft.nl}
% \and
}

\maketitle
\ificcvfinal\thispagestyle{empty}\fi

%%%%%%%%% ABSTRACT
% \begin{abstract}
%    The ABSTRACT is to be in fully-justified italicized text, at the top
%    of the left-hand column, below the author and affiliation
%    information. Use the word ``Abstract'' as the title, in 12-point
%    Times, boldface type, centered relative to the column, initially
%    capitalized. The abstract is to be in 10-point, single-spaced type.
%    Leave two blank lines after the Abstract, then begin the main text.
%    Look at previous ICCV abstracts to get a feel for style and length.
% \end{abstract}

%%%%%%%%% BODY TEXT
\section{Introduction}

% MSD poses the challenge to generate feasible room polygons based on an access graph while constraining possible layout by prexisting structural walls.
The Modified Swiss Dwellings (MSD) auto-completion challenge poses the problem of generating feasible room polygons based on an access graph while constraining possible layout by a structural walls mask. Previous work has tackled the problem of generating room polygons based on the access graph alone \cite{nauata2021housegan, shabani2022housediffusion}. In this report, it is experimented with if one of these previous works, HouseDiffusion \cite{shabani2022housediffusion}, can be adapted to this problem formulation by adding structural walls as an additional input.

% Difference in dataset complexity and number of training examples
In addition to conditioning on structural input, the Modified Swiss Dwellings dataset also brings other challenges. In the HouseDiffusion paper, the RPLAN dataset \cite{wu2019data} was used for training. RPLAN has around 60k training examples, and floor plans of a relatively small amount of rooms and corners. Modified Swiss Dwellings on the other hand has only around 5k training examples, with many more rooms per floor plan as well as more corners per room. In addition, while the room type is given in the RPLAN dataset, room type has to be predicted from a zoning type in MSD.

% Input of the MSD dataset
The input of the MSD dataset is given by 1) a segmentation mask of structural walls, and 2) a graph specifying the connections between rooms, and for each room a zoning type. As ground truth, MSD has both ground truth segmentation maps of room types and walls, as well as a graph representation. The graph representation has a node for each room, and the node attributes include the room geometry as well as the room type.

\section{Method}
\subsection{Modification to HouseDiffusion}

\paragraph{HouseDiffusion architecture} In HouseDiffusion each corner is represented by a corner embedding vector $C^t_{i, j}$ where $i$ is the index of the room, and $j$ the index of the corner within the room at time step $t$. The model denoises the floor plan by predicting an $x, y$ noise vector for each corner. The corner embedding vectors go through attention layers that contain 3 different types of masked attention: Component-wise Self Attention (CSA), Global Self Attention (GSA) and Relational Cross Attention (RCA). The type determines the attention mask that is used. The CSA attention is computed only between corners in the same room, GSA is computed between all corners in the floor plan, and RCA is computed between a corner and all corners in other rooms connected by a door.

\paragraph{Adding cross attention to structural corners} To add structure as an input constraint, the attention layers are modified by adding a cross-attention operation between each room corner and all structural corners. This is done by using the room corner embeddings as queries, and the structural corner embeddings as keys and values. The attention mask allows attention between each pair of room corner and structural corner.

\paragraph{Implementation details} 
% Before the structural corner embeddings are used in the structural cross attentions, they are encoded by a self-attention tranformer layers. 
The structural corner embeddings are passed through self-attention transformer layers, before being used in the cross attention between room corners and structural corners.
This is motivated by the hypothesis this might lead to more informative structural corner embeddings, as information about nearby structural corners can be used. No ablation study was performed to isolate the effect of adding this transformer.

Structural cross attention is added both in the continuous denoising steps, and in the discrete denoising steps.

% As a proof of concept this method was fisrst tried on the RPLAN dataset, where the outline of each floor plan was used as structural walls. See appendix \ref{app: RPLAN} to see results.

\subsection{Data preprocessing}

\subsubsection{Minimum Rotated Rectangle approximation}

% MSD has a lot more corners, thus is was chosen to use MRR approximation
In order to limit the number of corners of MSD floor plans, it was chosen to approximate the room polygons as minimum rotated rectangles (MRR). The rooms in MSD are not axis aligned, and thus minimum rotated rectangles form a better approximation than bounding boxes. An additional benefit of approximating rooms as MRRs is at inference time the number of corners for each room does not have to be sampled, but is always 4.

% Use of plotting smallest room last
At inference time, room polygons are formed by subtracting smaller rooms from larger rooms in cases of overlap. This is similar to a trick used by the rplanpy library \cite{wu2019data}, which represents RPLAN rooms by their bounding box and plots rooms ordered by largest area.

\begin{figure}
     \centering
     \begin{subfigure}[b]{0.3\linewidth}
         \centering
         \includegraphics[width=\linewidth]{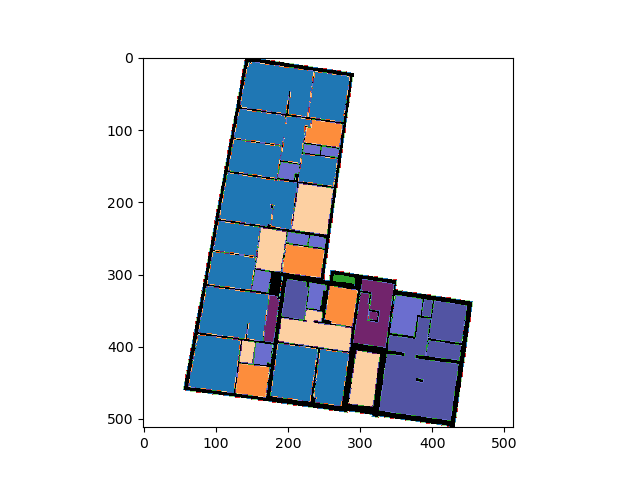}
         \caption{}
     \end{subfigure}
     % \hfill
     \begin{subfigure}[b]{0.3\linewidth}
         \centering
         \includegraphics[width=\linewidth]{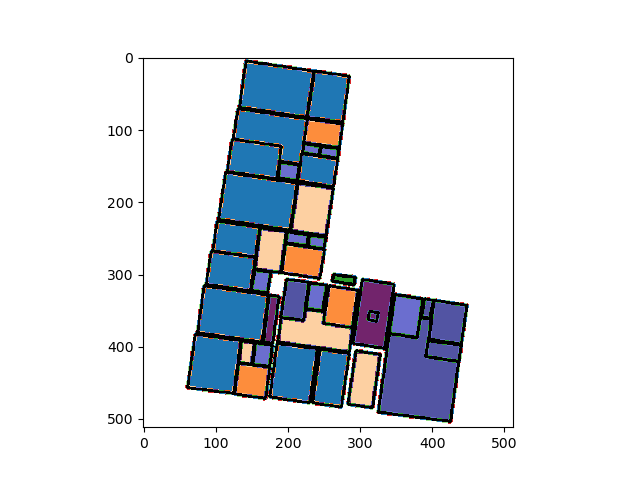}
         \caption{}
     \end{subfigure}
     % \hfill
     \begin{subfigure}[b]{0.3\linewidth}
         \centering
         \includegraphics[width=\linewidth]{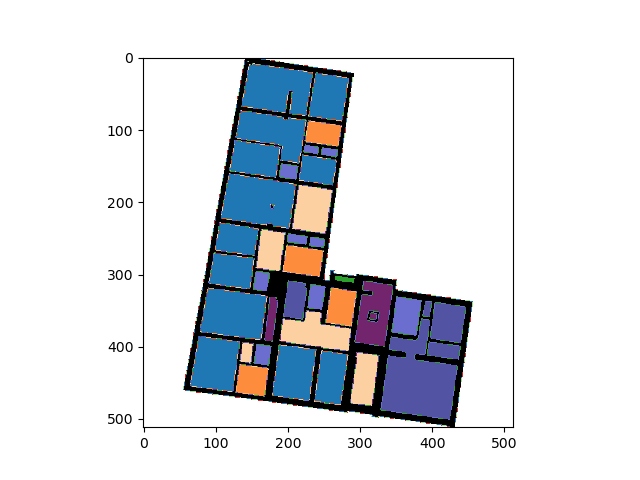}
         \caption{}
     \end{subfigure}
        \caption{(a) Original floor plan, (b) MRR approximation, and (c) MRR approximation refined by structural walls mask}
        \label{fig: mrr approximation}
\end{figure}

% Refine MRRs by structure
Because in MSD structural walls are part of the input, these can be used to further refine the minimum rotated rectangle approximations. If a rectangle is cut into multiple parts by a structural wall, only the largest part is used as room polygon. See figure \ref{fig: mrr approximation} for how well this approximation works on the training data. Table \ref{tab: results} shows a relatively high IoU (0.76) between the ground truth MRR approximations and the ground truth pixel-wise labels.

\subsubsection{Extracting structural walls from image}

In the dataset, the structural walls are given as segmentation mask images. However, the expected input for the modified HouseDiffusion model is a set of straight lines, each defined by two end points. A preprocessing step is thus needed to extract line endpoints for each wall segment.

The skeleton network library\cite{sknw} was used to extract a graph representation from a morphologically thinned version of the structural image. The lines output by this method were post-processed to 1) separate line segments into straight lines, and 2) filter out very short segments. The output of this preprocessing step is a collection of straight lines that approximate the wall structure image.

\subsubsection{Predicting room type from zoning type}

In the original HouseDiffusion model, room type was an input to the model. 
Instead of modifying HouseDiffusion to predict the room type, for simplicity it was chosen to predict the room type as a preprocessing step.

\paragraph{GNN Model}
PyTorch Geometric \cite{Fey/Lenssen/2019} was used to train a graph neural network to predict the room type for each node. The input to the model is a graph with one hot encoded zoning type as node features, and one hot encoded connection type (door, entrance, passage) as edge features. The architecture consists of $N$ GATConv \cite{veličković2018graph} graph convolutions. Each consecutive GATConv takes the same edge attributes as input. The output of the last GATConv is concatenated with the initial node features, and fed into a hidden linear layer. Then a final linear layer maps the hidden representation to the correct output dimension for predicting room type. Between each hidden layer, ReLU is used as an activation function, and dropout is applied.

\begin{table}[]
    \centering
    \begin{tabular}{rrr}
    \#GATConv layers & val loss & mean val acc \\
    2 & 0.35 & 0.87 \\
    3 & 0.30 & 0.90 \\
    4 & 0.35 & 0.87 \\
    8 & 0.41 & 0.83 \\
    16 & 0.58 & 0.74 \\
    \end{tabular}
    \caption{Hyperparameter search for number of GATConv layers}
    \label{tab: hyperparam search}
\end{table}

A hyperparameter search was done to determine a suitable number of GATConv layers $N = 3$. A learning rate of $10^{-3}$ and a batch size of $32$ were used. The models were trained with an early stopping tolerance of 5 for a maximum of 100 epochs. See table \ref{tab: hyperparam search} to see the results of the search. The accuracy is defined as the ratio of correctly predicted room types within a graph.

\begin{figure*}[t]
    \centering
    \includegraphics[width=\textwidth]{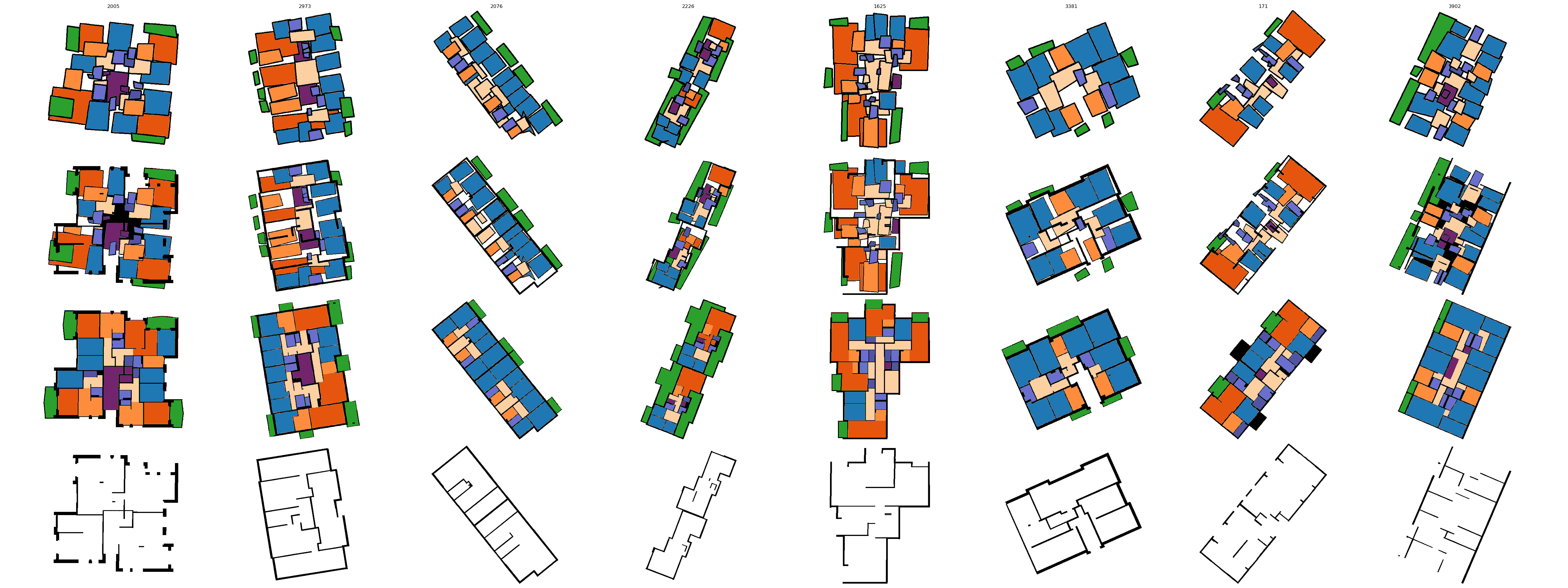}
    \caption{Example outputs on validation split. From top to bottom: predicted room rectangles, predicted room rectangles refined by structure mask, ground truth pixel-wise labels, structure masks.}
    \label{fig: val examples}
\end{figure*}

\begin{table*}[]
    \centering
    \begin{tabular}{rrrrrr}
    & w/o background & all & structure only & background only & w/o structure \\
    Refined by structure & 0.231 & 0.298 & 0.430 & 0.825 & 0.201 \\
    Predicted polygons only & 0.180 & 0.253 & 0.115 & 0.820 & 0.189\\
    \textit{GT MRR approximations} & \textit{0.737} & \textit{0.760} & \textit{0.373} & \textit{0.937} & \textit{0.791} \\

    \end{tabular}
    
    \caption{Validation set IoUs for the classes without background, all classes, structure only, background only, and w/o structure. \textit{Refined by structured} is the same model as submitted to CodaLab; \textit{predicted polygons only} are obtained without structural wall refinement; \textit{GT MRR approximations} shows the IoU between the ground truth MRR approximation, and the ground truth pixel-wise labels.}

    \label{tab: results}
\end{table*}

\section{Results \& Discussion}

For training, the MSD dataset was split into a train set of 3804, and a validation set of 100 samples. The rest of the samples (around 1000) were discarded for having too many rooms, to lower the amount of GPU memory needed for training. The model was trained for around 340k steps using a batch size of 32. During the last 100k steps, the loss did not improve significantly.

\paragraph{Qualitative results}

Example predictions on the validation set are shown in figure \ref{fig: val examples}. The top two rows show the predicted minimum rotated rectangle room polygons, MRR polygons refined by the structure masks respectively. The third row shows the ground truth pixel-wise labels, and the last row the structure masks.

The results show that the predicted room polygons follow the walls quite well, but are not very precise. Especially in cases of many smaller rooms, the rooms do not align together very well. The room polygons are also often not well enough aligned with the structural walls for the structure refinement post-processing step to work well.

The layout of the different room categories also visually looks similar between the ground truth and the predictions. However, to conclusively say if the model follows the room access graph well, a more in depth analysis is needed.

\paragraph{Quantitative results}

Similar evaluation to the CodaLab page are shown in table \ref{tab: results}. The table shows that the refinement by structure mask post-processing step has a positive effect on the IoU metrics, not only for the structure IoU, but on all classes.

The difference between the IoUs calculated on the validation versus those reported on CodaLab for the test set are small for the \textit{w/o background}: 0.231 vs 0.223, and \textit{structure only}: 0.430 vs 0.406. However, the IoU reported on CodaLab as \textit{w/ background} is most similar to the IoU calculated as \textit{background only}, even though it would be expected to be most similar to \textit{all}.

\paragraph{Future work \& limitations} This report shows that HouseDiffusion can be modified to consider structural walls as input. However, future work is needed to improve the quality of the results. Aspects that can be looked at include: is it useful to approximate the room polygons as minimum rotated rectangles, or is it better to use the original geometries? Are there better ways to pre-process and encode the structural walls? The amount of training data might also be too small for the problem complexity. This report also did not include ablations for adding a transformer encoder layer for the structural walls, and the effect of including the structural attention both in the continuous and discrete denoising steps.

{\small
\bibliographystyle{ieee_fullname}
\bibliography{egbib}
}

\end{document}